\documentclass[11pt,a4paper]{article}
\usepackage{times,latexsym}
\usepackage{url}
\usepackage[T1]{fontenc}

\usepackage[acceptedWithA]{tacl2021v1}

\usepackage{tacl2021v1}

\usepackage{xspace,mfirstuc,tabulary}

\newif\iftaclinstructions
\taclinstructionsfalse %
\iftaclinstructions
\renewcommand{\confidential}{}
\renewcommand{\anonsubtext}{(No author info supplied here, for consistency with
TACL-submission anonymization requirements)}
\newcommand{\instr}
\fi

\iftaclpubformat %

\else

\fi

\usepackage{times}
\usepackage{latexsym}
\usepackage{booktabs}
\usepackage{adjustbox}
\usepackage{microtype}
\usepackage{multirow}
\usepackage{amsmath}
\usepackage{caption}
\usepackage{inconsolata}
\usepackage{xcolor}       %
\usepackage{fontawesome5} %

\usepackage{wrapfig}

\usepackage{lineno}

\usepackage{listings}
\usepackage{xspace}
\usepackage{caption} 
\usepackage{pifont}

\usepackage{tcolorbox}

\usepackage{fancyhdr} 

\tcbuselibrary{listings,breakable,skins}

\definecolor{darkblue}{rgb}{0, 0, 0.5}
\hypersetup{colorlinks=true, citecolor=darkblue, linkcolor=darkblue, urlcolor=darkblue}

\newcommand\dataset{\textsc{Conflicts}}

\title{DRAGged into \dataset{}: \\Detecting and Addressing Conflicting Sources in Search-Augmented LLMs}

\newcommand{\bResearch}{\ding{171}}
\newcommand{\bbiu}{$\mathbin{\Diamond}$}

\author{Arie Cattan\textsuperscript{\bResearch{}\bbiu{}} \quad 
Alon Jacovi\textsuperscript{\bResearch{}} \quad 
Ori Ram\textsuperscript{\bResearch{}} \quad
Jonathan Herzig\textsuperscript{\bResearch{}} \quad 
\textbf{Roee Aharoni}\textsuperscript{\bResearch{}} \quad \\
\textbf{Sasha Goldshtein}\textsuperscript{\bResearch{}} \quad 
\textbf{Eran Ofek}\textsuperscript{\bResearch{}} \quad 
\textbf{Idan Szpektor}\textsuperscript{\bResearch{}}\quad 
\textbf{Avi Caciularu}\textsuperscript{\bResearch{}} \\
\textsuperscript{\bResearch{}}Google Research \quad \textsuperscript{\bbiu{}}Bar-Ilan University\\ 
\texttt{cattana@google.com} \\ 
}

\date{}

\begin{document}

\maketitle

\begin{abstract}

Retrieval Augmented Generation (RAG) is a commonly used approach for enhancing large language models (LLMs) with relevant and up-to-date information. However, the retrieved sources can often contain conflicting information and it remains unclear how models should address such discrepancies. 
In this work, we first propose a novel taxonomy of knowledge conflict types in RAG, along with the desired model behavior for each type.
We then introduce \dataset{}, a high-quality benchmark with expert annotations of conflict types in a realistic RAG setting. \dataset{} is the first benchmark that enables tracking progress on how models address a wide range of knowledge conflicts. 
We conduct extensive experiments on this benchmark, showing that LLMs often struggle to appropriately resolve conflicts between sources. While prompting LLMs to explicitly reason about the potential conflict in the retrieved documents significantly improves the quality and appropriateness of their responses, substantial room for improvement in future research remains.\footnote{Our dataset can be found at: \url{https://github.com/google-research-datasets/rag_conflicts}}

\end{abstract}

\section{Introduction}

\begin{figure}[t!]
    \centering
    \includegraphics[width=0.48\textwidth]{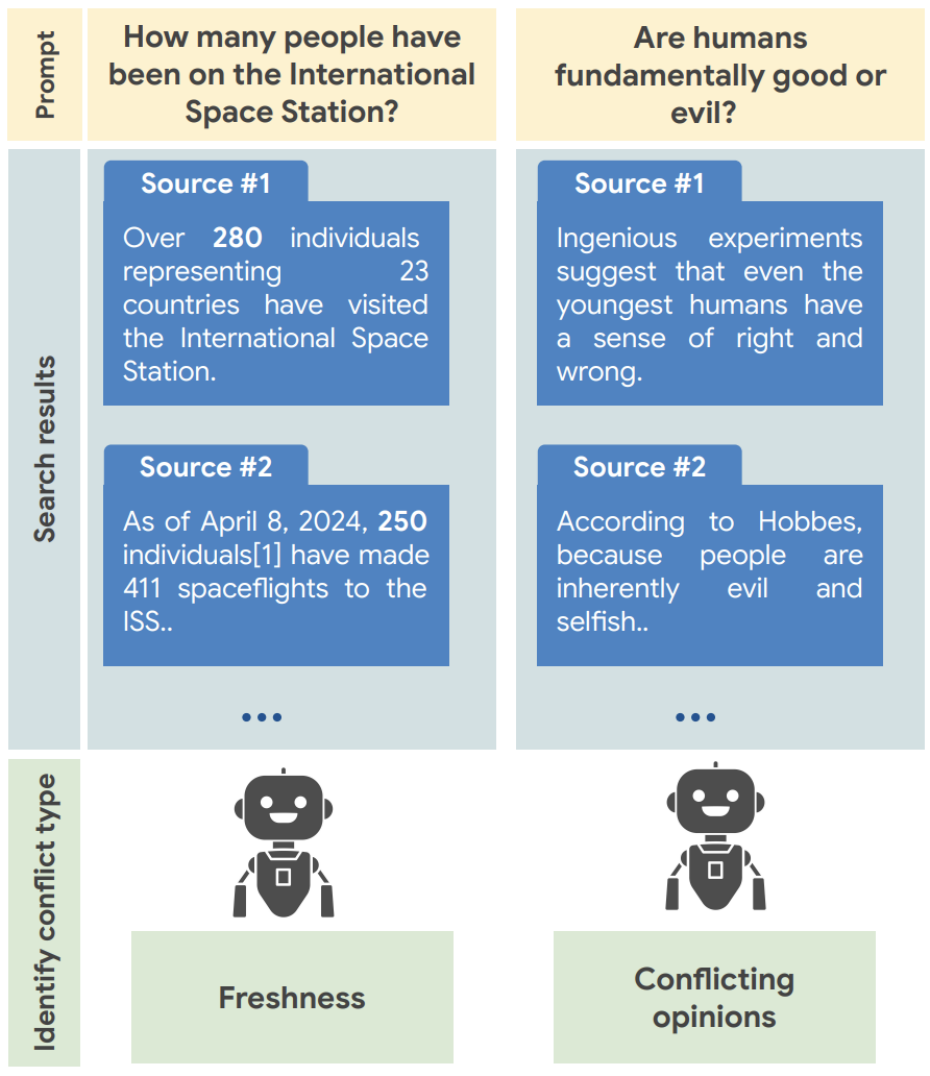}
    \caption{Examples of user queries with knowledge conflicts in the search results. 
    The left panel illustrates a \textit{freshness} conflict arising from outdated information, while the right panel shows \textit{conflicting opinions} on a controversial query. Identifying the conflict type is essential for generating the appropriate response (e.g., prioritizing recent data or presenting diverse perspectives).}
    \label{fig:taxonomy}
\end{figure}

Retrieval Augmented Generation~(RAG)~\citep{rag2020, realm} has emerged as an increasingly popular approach for improving the factual accuracy and relevance of large language model (LLM) outputs by leveraging external sources retrieved at inference time~\citep{Li2024EnhancingLF}. State-of-the-art commercial LLMs such as ChatGPT, Gemini and Claude have largely adopted this paradigm and developed ``search'' modes that retrieve up-to-date web content.

However, retrieved sources may provide conflicting information about the query and how models should optimally address these discrepancies continues to be an open and pressing research challenge~\citep{Xu2024DebateQAEQ, hou2024wikicontradict, Liu2024OpenDQ}. 
Prior research on this topic has typically focused on a specific type of conflict and proposed dedicated prompting strategies to address the conflict. For instance, \citet{Xu2024DebateQAEQ, hou2024wikicontradict} focused on debatable questions and prompted the model to explicitly highlight the conflict, \citet{Liu2024OpenDQ} assumed there is a single correct answer, and \citet{pmlr-v162-liska22a, vu-etal-2024-freshllms} focused on extracting up-to-date information. However, assuming that the the type of conflict is known in advance is often unrealistic in practice, and applying a specific method to different kinds of conflicts is sub-optimal.

In this work, we argue that knowledge conflicts arise for diverse reasons and should be addressed differently depending on the \emph{type} of conflict. 
Figure~\ref{fig:taxonomy} shows a few examples of queries with knowledge conflicts in their search results. Some discrepancies stem from temporal shifts (e.g., \emph{``How many people have been on the International Space Station?''}), where models should prioritize recent sources. Others reflect subjective opinions (e.g., \emph{``Are humans fundamentally good or evil?''}), where models should neutrally present the diversity of all the authoritative perspectives in the sources.

Inspired by these observations, we propose a comprehensive framework for handling knowledge conflicts. We introduce a novel taxonomy of conflict categories~(\S\ref{sec:taxonomy}; Table~\ref{tab:taxonomy}) and, for each category, define an \emph{expected model behavior} — a desired response style designed to emulate how a human would typically address that type of conflict.

To foster research on knowledge conflict resolution, we introduce \dataset{} (\S\ref{sec:dataset}), the first evaluation benchmark with explicit annotations of conflict types. To construct \dataset{}, we collected a total of 458 queries from various datasets known to contain diverse knowledge conflict types~\citep{vu-etal-2024-freshllms, Liu2024OpenDQ, zhang-choi-2021-situatedqa, wan-etal-2024-evidence}. For each query, we then retrieved a set of relevant source documents using Google Search. Finally, we tasked expert annotators with identifying whether the retrieved sources contained conflicting information and determining the conflict type according to our taxonomy. 
This annotation task is challenging because annotators need to comprehend a large volume of text to determine the appropriate conflict type. To ensure high-quality, we employ a three-stage annotation strategy: two expert annotators independently identify conflicts and their types, resolve disagreements through discussion, and a third expert annotator performs a final review. 
\dataset{} is the first benchmark to enable investigation into how well can models predict the conflict type and evaluate whether their outputs align with how humans would typically address that specific type of conflict.

\begin{table*}[t!]
\centering
\footnotesize

\begin{tabular}{p{0.2\textwidth}p{0.2\textwidth}p{0.5\textwidth}}
\toprule
\textbf{Category} &\textbf{Query} &\textbf{Retrieved Sources} \\
\midrule
No conflict 

& When did the Titanic set sail?
& [1] \textbf{\textit{On Wednesday 10th April 1912 shortly after 12noon}}, RMS Titanic set sail from Southampton’s White Star Dock on her maiden voyage to New York. \newline 
[2] On \textbf{\textit{April 1912}}, the Titanic set sail on its maiden voyage, traveling from Southampton, England, to New York City.\newline 
[3] On 11th April 1912 at 11.30am RMS Titanic dropped anchor in Queenstown, Ireland at Roches Point outer anchorage. \\

\midrule
Complementary Information & Is public transportation faster than driving in cities? & 
[1] Many areas have lanes dedicated to buses or high occupancy vehicles, which might \textit{\textbf{make taking a bus faster than driving yourself}}. \newline
[2] Even with worsening traffic, \textit{\textbf{driving still gets people to work faster}}, twice as fast in the U.S., a study by Governing found. \\
\midrule
Conflicting opinions or research outcomes & Is fasting beneficial for individuals with diabetes? & [1] \textit{\textbf{Intermittent fasting}}, when undertaken for health reasons in patients with diabetes mellitus, both types 1 and 2, \textit{\textbf{has been shown in a few small human studies to induce weight loss and reduce insulin requirements.}} \newline 
[2] This diet \textit{\textbf{is not recommended for those individuals with diabetes}}, children, the underweight, or with eating disorders, pregnant, or with chronic illnesses \newline
[3] Current evidence suggests that intermittent fasting \textit{\textbf{is an effective non-medicinal treatment option for type 2 diabetes. }}
\\
\midrule
Conflict due to outdated information & How many countries have recognized same-sex marriage? & [1] There are currently \textbf{\textit{37 countries}} where same-sex marriage is legal: Andorra ...  \newline 
[2] Same-sex marriage is legal in only \textbf{\textit{38 countries.}} \newline
[3] By 2022, same-sex marriage was legal in \textbf{\textit{32 countries.}} Since then, 3 more countries have joined this group: Andorra, Estonia, and Greece — bringing \textbf{\textit{the total to 35}}.  \\
\midrule 
Misinformation & When did season 5 of prison break come out? & [1] The season premiered on \textit{\textbf{April 4, 2017}}, and concluded on May 30, 2017, consisting of 9 episodes. \newline [2] Season 5 of the series was released on \textit{\textbf{May 30, 2017}}. It was filmed primarily in Dallas, Panama City and Los Angeles
\\

\bottomrule
\end{tabular}
\caption{
Examples from \dataset{} demonstrating each category of the knowledge conflict taxonomy. Each instance shows a query and a selection of corresponding retrieved passages.}
\label{tab:taxonomy}
\end{table*}

Finally, we conduct extensive experiments on \dataset{} using both open-source and commercial LLMs (\S\ref{sec:experiments_results}). Our results demonstrate that, while LLMs can generally produce accurate and factually grounded responses, they often fail to align with the expected type-specific behavior. Furthermore, we show that prompting models to explicitly reason about the potential conflict type substantially improves their performance. Despite these improvements, there remains a substantial room for improvement in future work.

Altogether, our work is the first to propose a \textit{general} approach for addressing a wide range of knowledge conflicts. We believe that our taxonomy and dataset will serve as a valuable resource for evaluating and improving future RAG models.

\section{Taxonomy of Knowledge Conflicts}
\label{sec:taxonomy}

We posit that the \textit{type} of knowledge conflict is important because different types induce different desired model behaviors. Consider, for example, the controversial query, \emph{``Is unlimited vacation time beneficial for employees?''}~\citep{wan-etal-2024-evidence}. Retrieved sources may present diverse viewpoints, including arguments for and against unlimited vacation time, as well as nuanced analyses of its pros and cons. In this scenario, a human would typically synthesize these arguments, presenting a balanced summary of the various perspectives. In contrast, some conflicts arise from temporal discrepancies--when  correct answers evolve over time. For instance, when faced with a query such as, \emph{``How many exoplanets have been discovered?''}, the retrieval process might yield a mix of up-to-date articles along with outdated ones, inevitably leading to numerical discrepancies. In this instance, human judgment would prioritize the most recent information.

To address these differences,
we introduce a taxonomy of conflict categories, denoted as $\mathcal{T}$. For each category $t$, we propose a corresponding \textit{expected behavior} $s = f(t)$, which defines the response style that approximates how humans might typically address queries from that type.
Below, we define the different types in our taxonomy along with the associated expected behavior.
Table~\ref{tab:taxonomy} presents examples for each category. Table~\ref{tab:expected_behavior_examples} illustrates the expected behavior of LLM outputs for each category.

\paragraph{No conflict} 
The retrieved documents provide answers to the query that are equivalent or nearly equivalent, referring to the same real-world entity, fact, or concept. Minor variations, such as differences in surface presentation or level of granularity (e.g., ``Wednesday 10th April 1912'' vs. ``April 1912'') do not constitute a conflict. Additionally, retrieved documents that are topically related to the query but do not directly answer it, are not considered sources of conflict. 
For example, given the query ``\textit{When did the Titanic set sail?}'' in Table~\ref{tab:taxonomy}, the search result [3] mentioning the drop anchor in Queenstown on 11th April is related but does not answer the query. The expected behavior for this category is to provide a clear and direct answer without introducing alternative viewpoints or uncertainty~\citep{yona-etal-2024-large}.

\paragraph{Complementary Information}
The retrieved documents provide answers to the query that refer to \textit{different} real-world concepts but are mutually compatible, meaning that a single person can reasonably agree with all of them simultaneously. This can happen when there are multiple correct answers to the same query (e.g, ``\textit{ What is the meaning of CI in police?}'' has two valid answers: Confidential Informant or Circle Inspector) or when the query is underspecified and can be answered with multiple complementary perspectives (e.g., the question ``\textit{Is public transportation faster than driving in cities?}'' which depends on the city). The expected behavior for this category is to consolidate and reconcile the different partial answers provided by the retrieved documents, without framing the response as a debate~\citep{stelmakh-etal-2022-asqa}.\footnote{A few works address underspecified queries by generating clarifying questions~\citep{Aliannejadi2019AskingCQ, zhang-choi-2025-clarify}. However, in this work, we aim to propose a general approach for addressing the diverse range of conflicts and therefore opt for generating a consolidated response.}

\paragraph{Conflicting opinions or research outcomes} 
The retrieved documents provide answers to the query that are \textit{not} mutually compatible. This category includes subjective queries that elicit conflicting opinions (e.g., \emph{``Are humans fundamentally good or evil?''}), queries with contradictory research findings (e.g., \emph{``Is paracetamol more effective than placebo?''}), lack of historical consensus, etc., where the retrieved sources disagree and argue towards a specific side.\footnote{We focus exclusively on safe queries and leave the identification of harmful or hateful content for future work.} 
The expected behavior for this category is to explicitly reflect the debate between the retrieved sources and to neutrally summarize the different viewpoints~\citep{hou2024wikicontradict, slobodkin-etal-2024-multi, Xu2024DebateQAEQ}.

\paragraph{Conflict Due to Outdated Information\footnote{Hereafter, we use the term ``Freshness'' to refer to this conflict type for brevity.}}
The retrieved documents provide answers to the query that are \textit{not} mutually compatible, but the conflict stems from temporal discrepancies—some sources reflect outdated information, while others provide more recent updates. For example, given the query ``\textit{How many countries have recognized same-sex marriage?}'' in Table~\ref{tab:taxonomy}, the retrieved sources report 37, 38, or 35, depending on their publication dates. The expected behavior is to prioritize the up-to-date information~\citep{vu-etal-2024-freshllms}, while optionally acknowledging the presence of outdated sources.

\paragraph{Conflict Due to Misinformation} 
The retrieved documents provide answers that are \textit{not} mutually compatible, where at least one source contains information that is likely false, misleading, or inaccurate. For example, in response to the query \emph{``When did season 5 of prison break come out?''}~(Table~\ref{tab:taxonomy}), one source inaccurately states \textit{``May 30, 2017''} instead of \textit{``April 4, 2017''}. The expected behavior for this category is to disregard inaccurate sources and to provide a response grounded in reliable and verified information~\citep{pan-etal-2023-attacking, jiayang-etal-2024-econ, ming2025faitheval}.

\section{\dataset{}}
\label{sec:dataset}

This section introduces \dataset{}, the first dataset annotated with knowledge conflict types. Each instance within \dataset{} 
comprises a query, a set of retrieved relevant passages, a corresponding conflict type label, and, for specific types (detailed below), the ground truth correct answer.

\paragraph{Queries and retrieved documents}

To ensure coverage across different types of knowledge conflicts, we curated seed queries from several existing datasets. We first include queries from FreshQA~\citep{vu-etal-2024-freshllms}, which focuses on questions requiring fast-changing world knowledge; SituatedQA Temp and SituatedQA Geo~\citep{zhang-choi-2021-situatedqa} that include underspecified queries whose answers depend on temporal and geographic context, respectively; and QACC~\citep{Liu2024OpenDQ}, which provides unambiguous queries.
For these datasets, we use Google Search to retrieve the top-10 search results for each query. This process yields document titles, short snippets, and where available, publication dates. We also consider the recent ConflictingQA dataset~\citep{wan-etal-2024-evidence}, where each instance consists of a Yes/No query paired with supporting evidence for both sides, originally retrieved using Google Search.

We find that the automatically generated Google Search snippets often omit crucial context for resolving knowledge conflicts and generating appropriate answers. For example, the query \emph{``When did Toyota first come?''} yields different snippets that mention seemingly contradictory dates (i.e., \textit{1933}, \textit{1937}, \textit{1955}, etc.). However, analyzing the full articles reveals that these discrepancies stem from references to distinct aspects of Toyota's history (e.g., different company divisions, initial car releases, etc.).
Therefore, we parse the complete text from the HTML pages using cloudscraper\footnote{\url{https://github.com/VeNoMouS/cloudscraper}} and jusText\footnote{\url{https://github.com/miso-belica/jusText}}. 
Following \citep{wan-etal-2024-evidence}, we then extract the most relevant 512-token segments from each document by applying the TAS-B model \citep{retrieval2021} across overlapping 512-token windows with a 256-token stride, and calculate the dot product between the window's embedding and the query's embedding.

\paragraph{Annotation Process}
We cannot automatically annotate the conflict type based solely on the data sources (e.g., queries from FreshQA defaulting to Frehsness) because FreshQA, QACC and SituatedQA are only a collection of queries without relevant documents. These queries are \textit{susceptible} to lead to knowledge conflicts between some sources, but do not necessarily guarantee conflicts among the top-10 search results. Although ConflictingQA includes inter-context annotation of knowledge conflict for Yes/No questions, it does not differentiate between the nuanced types of knowledge conflict of our taxonomy (e.g., \textit{Complementary information} vs. \textit{conflicting opinions}).

Therefore, we turn to human annotation by presenting annotators with the queries and their corresponding webpage segments, instructing them to label the conflict type based on our taxonomy (Section \ref{sec:taxonomy}). During the annotation, we add two additional categories: \textit{``Other''} to capture any conflict that does not correspond to one of the defined categories in our taxonomy and \textit{``No relevant sources''} if Google Search yielded no relevant results. In practice, no instances were annotated as ``Other'', suggesting that our taxonomy is comprehensive and captures well the different conflict types. 
In addition to the conflict type, annotators were asked to provide brief, unstructured explanations for their decisions. In cases of ``No conflict'', ``Conflict due to outdated information'' and ``Misinformation'', annotators were also required to write the correct and up-to-date answer from the search results.

This annotation task poses a considerable challenge, as it requires annotators to carefully examine all search results to accurately identify the type of knowledge conflict in each instance and, when applicable, select the correct answer.
Moreover, the boundaries between some categories are often subtle and require a deep understanding of the search results. For example, identifying whether different search results complement each other or present conflicting viewpoints requires annotators to assess whether the same person could plausibly agree with all points of view. This distinction is important as it affects the expected style of the response: complementary information calls for cohesive aggregation, while conflicting viewpoints require a neutral summary highlighting disagreements (as discussed in Section~\ref{sec:taxonomy}). Additionally, annotators must disregard search results that are irrelevant to the query.

To ensure high-quality annotations, each instance was annotated by two independent annotators with linguistic expertise, followed by a reconciliation phase to resolve discrepancies and a review step performed by a third expert annotator. 
To facilitate the annotation process, we automatically generate a response for each separate search result using Gemini Pro 1.5~\citep{Reid2024Gemini1U} and present them to the annotators. These responses are then used for easily identifying whether the search results provide equivalent or different information about the query without reading all search results. For example, given the query \emph{``What was the religion in Persia before Islam?''}, Gemini generates the same answer (``\textit{Zoroastrianism}'') for each separate source. However, the annotators were explicitly instructed to treat those model-generated responses as optional hints and to disregard them if they appear to be hallucinations.

\begin{table}[t]
    \centering
    \footnotesize
    \begin{tabular}{lr}
    \toprule
         Conflict Type & \# Instances \\
         \midrule
         No conflict & 161 \\
         Complementary information & 115 \\ 
         Conflicting opinions & 115\\
         Outdated Information & 62\\
         Misinformation & 5\\
         \midrule
         Total & 458 \\
         \bottomrule
    \end{tabular}
    \caption{Statistics of \dataset{}.}
    \label{tab:data_stats}
\end{table}

\paragraph{Statistics}
The final \dataset{} dataset contains 458 instances after filtering out 18 instances with \textit{No relevant sources}. There are an average of 9.2 search results for each query. Table~\ref{tab:data_stats} presents the distribution of conflict types in \dataset{}.

65\% of the instances in \dataset{} (297 of 458) were flagged as conflicting (\textit{Complementary Information}, \textit{Conflicting opinions}, \textit{Freshness}, and \textit{Misinformation}) and thus require reasoning and aggregating spread information across multiple sources in order to produce an appropriate response. Annotators identified only 5 cases of \textit{Misinformation}, likely because genuine misinformation is rare ``in the wild'', especially among the top-10 search results, where modern search engines are optimized to demote or blocklist low-quality and misleading content. Indeed, most previous work on misinformation has automatically perturbed text to simulate misinformation~\citep{Du2022SyntheticDA, pan-etal-2023-attacking, jiayang-etal-2024-econ, ming2025faitheval}.

The remaining 161 instances are those annotated as ``No conflict''. Although lacking explicit knowledge conflicts, these instances present other common RAG challenges such as handling long context, dealing with irrelevant search results and resolving ambiguity. For example, given the query \textit{``Where do peaches come from?''}, some search results properly mention that Peaches originated in China, while other sources state personal opinions about where to find good peaches nowadays \emph{``the best peaches I have ever put in my mouth come from Goldbud Farms in Placerville, in the middle of California GoldRush country''}. In such cases, models should prioritize the search results identifying the origin of peaches and disregard the latter, which do not answer the query.

It is important to note that the conflict type distribution in \dataset{} is a result of our curated query selection and do not necessarily reflect the natural distribution of conflicts in search engines.

\section{Tasks}
\label{sec:tasks}

The taxonomy~(\S\ref{sec:taxonomy}) and dataset~(\S\ref{sec:dataset}) enables the investigation of new research questions on RAG with knowledge conflicts. We can explore whether models can accurately predict the type of conflict, assess how well model outputs align with the expected response behavior, and investigate whether explicitly leveraging the conflict type can improve response quality. To address these questions, we formalize two core tasks:

\noindent \paragraph{\textit{Task 1 (Classification)}: Conflict type prediction.} Given a query $q$ and the retrieved relevant paragraphs $\mathcal{C}_q$, the goal is to predict the category of the knowledge conflict $\hat{t}$ from the taxonomy between the retrieved documents $\mathcal{C}_q$ with respect to the query $q$, as follows: 
\begin{equation}\label{eq:conflict}
    \hat{t} \sim p_{\theta}(t | [q; \mathcal{C}_q])
\end{equation}

This auxiliary task can guide downstream response generation (\S\ref{sec:experiments_results}) and enable the evaluation of models' understanding of the underlying relationships between the different sources.

\noindent \paragraph{\textit{Task 2 (Generation)}: Generating an appropriate response.} Given a query $q$ and the retrieved relevant paragraphs $\mathcal{C}_q$, the goal is to generate a grounded and accurate response $\hat{y}$ that conforms to the expected behavior:
\begin{equation}\label{eq:qa}
    \hat{y} \sim p_{\theta}(y | [q;\mathcal{C}_q]) 
\end{equation}

\begin{table*}[t!]
\centering
\resizebox{0.98\textwidth}{!}{
\footnotesize
\setlength{\tabcolsep}{4pt} %

\begin{tabular}{m{0.12\textwidth}m{0.12\textwidth}m{0.4\textwidth}m{0.4\textwidth}}

\toprule
\textbf{Query} &\textbf{Conflict type} &\textbf{Responses} &\textbf{Comment} \\
\midrule
\multirow{2}{\hsize}{When did season 5 of prison break come out?} &\multirow{2}{\hsize}{Misinformation} &Season 5 of Prison Break premiered on April 4, 2017, on Fox [3]. The season was released on May 30, 2017 [4]. & \textcolor{red}{\faTimes} Although the model includes the correct answer (April 7, 2020), the style of the response is incorrect because the model tries to reconcile between the sources, while the source [4] is incorrect. \\
\cmidrule{3-4}
& &Season 5 of Prison Break premiered on April 4, 2017 [2][3]. The first episode aired on this date [2]. & \textcolor{green}{\faCheck} The response provides a clear and coherent answer to the question. \\
\midrule
\multirow{2}{\hsize}{How many people have been on the International Space Station?}
& \multirow{2}{\hsize}{Freshness} &

As of May 2022, 258 individuals from 20 countries have visited the International Space Station (ISS) [5]. Other sources state over 260 individuals from 20 countries and five International Partners have visited the station [7]. 
& \textcolor{red}{\faTimes} The response presents conflicting opinions about the query, whereas source [5] is outdated. \\
\cmidrule{3-4}
 && As of March 2024, 279 individuals from 22 countries have visited the International Space Station [2]. Older figures indicate that as of May 2022, 258 people from 20 countries had visited the station [5], and as of September 12, 2021, 244 people had made 403 spaceflights to the ISS [3]. 
& \textcolor{green}{\faCheck} The response clearly answers the question, stating there were 279 people as of March 2024, and acknowledges older figures from other sources. \\
\midrule
\multirow{2}{\hsize}{Who has the biggest airport?} & \multirow{2}{\hsize}{Complementary information} &
The King Fahd International Airport in Saudi Arabia is the largest airport in the world by land area [1][6][7]. It spans nearly 300 square miles, an area comparable to the size of New York City's five boroughs combined [6]. 
& \textcolor{red}{\faTimes} The model writes the response only with the perspective of the largest airport. \\
\cmidrule{3-4}
&& The world's largest airport can be defined in several ways, including by land area, passenger volume, or flight volume [6]. King Fahd International Airport in Saudi Arabia is the largest airport by total area [1]. In terms of passenger traffic, Hartsfield-Jackson Atlanta International Airport in the United States is the busiest airport in the world [4, 5]. 
& \textcolor{green}{\faCheck} The response presents multiple aspects of the what is considered a ``big'' airport, in terms of land area and passenger volume. \\
\midrule
\multirow{2}{\hsize}{Where did the style hot dog originated?} &
\multirow{2}{\hsize}{Conflicting opinions} & 
The Chicago-style hot dog originated during the Great Depression in the 1930s in Chicago [1]. It was created to provide a hearty and affordable meal, with a nickel buying a hot dog loaded with condiments [1]. 
& \textcolor{red}{\faTimes} The model only writes the origin of a specific hot-dog (Chicago-style). \\
\cmidrule{3-4}
&& The exact origin of the hot dog is not definitively clear, making it a subject of some debate among historians [6]. Two European cities, Frankfurt, Germany, and Vienna, Austria, both claim to be the birthplace of the sausage that served as the precursor to the hot dog [7]. The Chicago-style hot dog, with its specific set of toppings, emerged later, during the Great Depression in the 1930s [1]. 
& \textcolor{green}{\faCheck} The response presents a debate around the origin of hot-dog. \\

\bottomrule
\end{tabular}}
\caption{Examples of model responses that adhere and do not adhere the expected behavior for each category in our taxonomy.}
\label{tab:expected_behavior_examples}

\end{table*}

\paragraph{Evaluation.} We conduct a multi-faceted evaluation to assess the quality of the generated response $\hat{y}$, including factual grounding, accuracy (where applicable), and adherence to the expected behavior. 

First, following common practices in grounded generation~\citep{Jacovi2025TheFG}, we evaluate the \textbf{factual grounding} of the response $\hat{y}$ with respect to the retrieved search results. Specifically, since we require models to generate grounded responses with inline citations for each sentence pointing to the relevant search results (see Section~\ref{subsec:experiments}), we measure citation quality~\citep{gao-etal-2023-enabling, slobodkin-etal-2024-attribute}. 
We adapt the prompt from the FACTS benchmark~\citep{Jacovi2025TheFG}, asking the model to assess whether each sentence is ``supported'', ``unsupported'', ``contradictory'' or ``no factual information''. The factual grounding score of an entire response is the percentage of ``supported'' sentences over all sentences with factual information.

Second, for instances with a single correct answer, we evaluate whether the generated response $\hat{y}$ correctly incorporates the gold answer $y$ from \dataset{}. We refer to this evaluation as  \textbf{Answer Recall}. This applies to instances in the ``No conflict'', ``Freshness'', and ``Misinformation'' categories, which comprise 228 queries in \dataset{}. Following~\citep{mallen-etal-2023-trust, liu-etal-2024-lost}, we consider a response correct if the gold answer $y$ is included in $\hat{y}$. We avoid strict string matching because LLM outputs are often verbose or paraphrastic yet correct and instead consider an LLM-based evaluator to assess semantic inclusion of the gold answer.

Third, and specific to our task of resolving conflicting information, we assess whether the generated response $\hat{y}$ adheres to the \textbf{expected behavior} $s$ associated with the conflict type $t$, as defined in Section~\ref{sec:taxonomy}. 
Examples of model outputs, along with their evaluation for adherence to the expected behavior, are shown in Table~\ref{tab:expected_behavior_examples}.
This evaluation task requires assessing nuanced stylistic aspects of the response beyond factual consistency or answer inclusion and cannot be achieved with simple heuristics. For instance, in the query \textit{``How many people have been on the International Space Station?''} (Table~\ref{tab:expected_behavior_examples}), the first response fails to adhere to the expected behavior by framing the different figures as a debate between the sources. In contrast, the second response clearly distinguishes between older and newer answers (i.e., ``Older figures'', ``as of September 12, 2021'').

As LLMs have demonstrated remarkable evaluation capabilities across a range of tasks, sometimes matching expert human raters~\citep[\textit{inter alia}]{chiang-lee-2023-large, Zheng2023JudgingLW, liu-etal-2023-g, kocmi-federmann-2023-large, kamalloo-etal-2023-evaluating}, we employ an LLM for assessing adherence to the expected behavior. 
This evaluator is few-shot prompted with examples of adherent and non-adherent responses. Specifically, we design a separate prompt template for each conflict type, incorporating: (i) the query, (ii) the conflict type description and associated expected behavior, (iii) 2–3 positive and negative examples, and (iv) the candidate response. The model outputs a binary decision indicating whether the response adheres to the expected behavior.
We validate this automatic rater against human judgments on a subset of 100 examples from the dataset, achieving an accuracy of 0.89, which demonstrates the reliability of our automatic evaluator.

Together, these three evaluation metrics--factual consistency, answer recall, and expected behavior adherence--provide a comprehensive assessment of response quality, capturing different critical aspects. For instance, a response might be factually consistent with the search results and even include the gold answer but fail to adhere to the appropriate behavior for the conflict type, e.g., by providing an additional incorrect fact grounded to one of the search results (see the first response to the query ``When did season 5 of prison break come out?'' in Table~\ref{tab:expected_behavior_examples}). Conversely, a response could align perfectly with the expected behavior yet provide an incorrect answer or contain claims not supported by the sources. 

\section{Experimental Setup}
\label{sec:experiments_results}

\subsection{Experiments}
\label{subsec:experiments}

We conduct extensive experiments on \dataset{} to address our two core tasks (Section~\ref{sec:tasks}): (1) conflict type prediction, and (2) generation of a response that adheres to the appropriate behavior. For both tasks, each search result is represented as a concatenation of its URL, page title, Google snippet, publication date (if available), and the 512-token segment (\S\ref{sec:dataset}).

\begin{table}[!t]
    \centering
    \begin{tabular}{lc}
    \toprule
        Model &  Accuracy \\
        \midrule
        Gemma 3 27B & 53.9 \\
        Qwen 2.5 72B & 53.1 \\
        GPT-4o & 59.2 \\
        Gemini 2.0 Flash & 60.5 \\
        Gemini 2.5 Flash & \textbf{65.3} \\
        \bottomrule
    \end{tabular}
    \caption{Performance (accuracy) of models for predicting the conflict category on \dataset{}.}
    \label{tab:conflict_type_results}
\end{table}

\paragraph{Conflict type prediction:}
We prompt the model with the query, retrieved evidence, and our taxonomy $\mathcal{T}$ (Table~\ref{tab:taxonomy}), which includes category definitions and 1–2 illustrative examples per class. The model is then asked to classify the type of conflict

\paragraph{Response Generation:}
We explore multiple prompting strategies for generating the response $\hat{y}$:
\begin{enumerate}
    \item \textbf{Vanilla:} A standard RAG-style approach where the model receives the query and search results as input and generates a response: $p_{\theta}(y | [q; \mathcal{C}_q])$~\citep{ram-etal-2023-context}.
    \item \textbf{Pipeline}: A two-step process. First, the model predicts the conflict type $\hat{t}$ given the taxonomy $\mathcal{T}$, the query and its search results: $p_{\theta}(t | [\mathcal{T};q;\mathcal{C}_q])$. Second, the model generates the response $\hat{y}$ using the predicted conflict type $\hat{t}$ as additional context: $p_{\theta}(y | [q; \mathcal{C}_q; \hat{t}])$. 
    \item \textbf{Taxonomy Aware:} A joint approach where the model simultaneously predicts the conflict type and generates the response in a single pass. The prompt includes the full taxonomy $\mathcal{T}$ along with the query and search results: $p_{\theta}(t, y | [\mathcal{T}; q; \mathcal{C}_q])$. 
    \item \textbf{Oracle:} An upper-bound setting in which the model is given the gold conflict type $t^*$ from the dataset in addition to the query and search results: $p_{\theta}(y | [q; \mathcal{C}_q; t^*])$.
\end{enumerate}

For all prompts, we also instruct the model to provide in-line citations (e.g., [2]) to the relevant search results for each sentence~\citep{gao-etal-2023-enabling}.

\begin{table*}[!ht]
\centering

\resizebox{0.95\textwidth}{!}{
\begin{tabular}{llccc}\toprule
Model &Prompt &Expected Behavior &Answer Recall & Factual Grounding \\\midrule

\multirow{4}{*}{Gemma 3 27B~\citep{Kamath2025Gemma3T}} &Vanilla & 59.4 &89.0 &94.4 \\
&Pipeline & 71.6 &89.9 &91.9 \\
&Taxonomy-Aware & 76.2 & 86.0 & 92.9 \\
&Oracle & 87.6 &88.6 &93.3 \\
\midrule

\multirow{4}{*}{Qwen 2.5 72B~\citep{Yang2024Qwen25TR}} & Vanilla &64.2 &87.7 &90.5 \\ 
&Pipeline &67.5 & 89.0  & 88.2 \\
&Taxonomy-Aware & 68.3 & 88.2 & 89.4 \\
&Oracle &90.6 &87.3 &89.3  \\
\midrule
\multirow{4}{*}{GPT-4o~\citep{Hurst2024GPT4oSC}} & Vanilla  &67.2 &87.3 &91.2 \\
&Pipeline & 74.9 &87.3 &92.0 \\
&Taxonomy-Aware & 70.0 & 88.6 & 94.0\\
&Oracle  &88.6 &87.3 &91.4  \\
\midrule
\multirow{4}{*}{Gemini 2.0 Flash~\citep{google_gemini_2.0}} &Vanilla  & 62.0 & 90.8 & 98.5 \\
&Pipeline & 75.3 & 89.5 & 96.6 \\
&Taxonomy-Aware & 69.4 & 89.0 & 96.8 \\
&Oracle & 87.8 & 90.4 & 96.8 \\
\midrule
\multirow{4}{*}{Gemini 2.5 Flash~\citep{google_gemini_2.5}} &Vanilla & 65.7 & 92.1 & 96.8 \\
&Pipeline & \textbf75.1 & 90.8 & 96.9 \\
&Taxonomy-Aware & 69.9 & 94.3 & 96.2 \\
&Oracle &88.0 &91.7 &96.8 \\
\midrule
\multirow{4}{*}{Gemini 2.5 Flash Thinking} &Vanilla  &68.3 &90.8 &96.2 \\
&Pipeline & 76.6 &89.0 &96.1 \\
&Taxonomy-Aware &  73.6 & 91.7 & 97.0 \\
&Oracle &89.3 &89.9 &96.8 \\
\bottomrule
\end{tabular}}
\caption{Performance of response quality. For each model and prompt strategy, we report the expected behavior accuracy, answer recall and factual grounding~(\S\ref{sec:tasks}).}
\label{tab:results}
\end{table*}

\subsection{Results}

We present the results of the conflict type prediction in Table~\ref{tab:conflict_type_results}, with Gemini 2.5 Flash achieving the highest accuracy (65.3\%). 

Table~\ref{tab:results} presents the results of the response generation quality. For each model and prompt template, we report the accuracy of the expected behavior, answer recall and factual consistency, using Gemini 2.5 Flash~(\S\ref{sec:tasks}).

\paragraph{Result 1: Model responses exhibit limited adherence to the expected behavior.}
Table~\ref{tab:results} shows that the standard RAG prompt (Vanilla) generates responses that moderately adhere to the expected behavior, with scores ranging from 59.4 for the open-source Gemma 3 27B to 68.3 for Gemini 2.5 Flash with Thinking mode. These relatively low scores reveal that, while models can be grounded on the search results and include the correct answer, they sometimes fail to follow the expected behavior. This highlights the importance of evaluating not only the factual accuracy and grounding of model outputs, but also whether the \textit{style} of the response aligns with human preferences.

\paragraph{Result 2: Explicitly incorporating conflict type
improves expected behavior.} The Oracle prompt, which augments the LLM input with the gold conflict type from \dataset{}, substantially improves adherence to the expected response behavior across all models. On average, it yields a 24-point gain over the Vanilla prompt (e.g., +28.2 for Gemma, +21.4 for GPT-4o, +21.0 for Gemini 2.5 Flash with Thinking mode, etc.), while preserving high answer recall and factual consistency scores. These results indicate that models have the general capability to generate appropriate responses and there is considerable room in developing methods that can approximate this upper bound.

\paragraph{Result 3: Pipeline and Taxonomy-aware prompts improve the expected behavior.}

Table~\ref{tab:results} shows that both the pipeline and the taxonomy-aware prompts improve adherence to the expected behavior over the vanilla approach, without degrading the answer recall and factual grounding scores. On average, they yield performance gains of 9 and 5.5 points, respectively. This suggests that prompting models to reason explicitly on the potential knowledge conflict in the search results, can substantially improve response quality. 

Nonetheless, search augmented LLMs must address a range of additional challenges beyond knowledge conflicts, including handling irrelevant sources due to an imperfect retrieval \citep{yoran2024making}, determining when to refine the user query and to search for additional evidence, or addressing queries with safety concerns. Therefore, future work can explore how to integrate such conflict-aware methods into practical RAG systems.

\subsection{Analysis}
\label{subsec:analysis}

\paragraph{Expected behavior per category}

\begin{table*}[!htp]\centering
\footnotesize
\begin{tabular}{lcccc}\toprule
Category &Vanilla & Pipeline & Taxonomy-Aware &Oracle \\
\midrule
No conflict &78.4 & 74.7 & 71.0 & 90.1 \\
Complementary Information & 83.3 & 83.3 & 78.1 & 94.7 \\
Freshness and Misinformation & 74.2 & 75.8 &78.8 & 80.3 \\
Conflicting opinions & 36.2 & 73.3 & 69.8 & 87.9 \\
\bottomrule
\end{tabular}
\caption{Expected behavior accuracy of Gemini Flash 2.5 Thinking per category. We combine Freshness and Misinformation because both categories require selecting the correct response.}
\label{tab:results_per_category}
\end{table*}

Table~\ref{tab:results_per_category} presents the expected behavior evaluation for each category in our taxonomy~(\S\ref{sec:taxonomy}). For brevity, we report the performance of Gemini 2.5 Flash with thinking mode, other models showing similar trends. The most challenging category is \emph{``Conflicting opinions''} with Gemini achieving only 36.2\% under the Vanilla prompt. The pipeline approach improves the expected behavior for \textit{Complementary information}, \textit{Freshness and Misinformation} and \textit{Conflicting opinions}, with a slight drop in \textit{No conflict}.

\paragraph{Error Analysis}
To better understand the headroom in conforming to the expected behavior, we manually analyze 40 randomly sampled outputs that do not adhere to the expected behavior from our best-performing model (Gemini 2.5 Flash with thinking). For the \textit{Complementary information} category, the most common error was under-specification: model response often include only one correct answer, failing to capture the full range of relevant results (see Table~\ref{tab:expected_behavior_examples}). For \textit{Conflicting opinions}, models either present only a single viewpoint or multiple viewpoints with a strong bias toward one perspective. In the \textit{No conflict} and \textit{Freshness} categories, models frequently hedge by expressing uncertainty or mentioning multiple possible answers.

\section{Related Work}
\label{subsec:related_work}

\subsection{Retrieval Augmented Generation}
\label{subsec:ralm}

Retrieval Augmented Generation (RAG) consists of conditioning a model on relevant documents from a large corpus during generation~\citep{realm, rag2020,atlas, Borgeaud2021ImprovingLM, ram-etal-2023-context, Gao2023RetrievalAugmentedGF}. Retrieved documents can bring conflicting information, which can complicate the generation process. 
Previous work on knowledge conflicts has mostly focused on a single type of conflict, such as conflicts arising from outdated information~\citep{pmlr-v162-liska22a, kasai2023realtime, vu-etal-2024-freshllms}, controversial queries with disputed opinions~\citep{wan-etal-2024-evidence, Xu2024DebateQAEQ}, ambiguous queries~\citep{min-etal-2020-ambigqa, zhang-choi-2021-situatedqa, lee2024ambigdocs}, or factual contradictions observed in real-world scenarios~\citep{hou2024wikicontradict} or in synthetically created datasets~\citep{wang2024resolving, jiayang-etal-2024-econ, tan-etal-2024-blinded}. 
A related line of research explores context-memory conflicts, where retrieved documents contradict the model’s parametric knowledge~\citep{longpre-etal-2021-entity, kortukov2024studying}. For a broader survey of knowledge conflicts in LLMs, we refer the reader to~\citep{xu-etal-2024-knowledge-conflicts}.
This work focuses on inter-context knowledge conflicts and proposes a comprehensive taxonomy of conflict types for appropriately addressing the diverse range of conflicts.

\subsection{Datasets with Knowledge Conflicts}

In recent years, several QA datasets with knowledge conflicts between different sources were introduced. For example, ConflictingQA~\citep{wan-etal-2024-evidence} generates controversial queries with LLMs and automatically identifies the stance (Yes or No) of each search result. Similarly, DebateQA~\citep{Xu2024DebateQAEQ} collects many debatable questions from various sources and automatically generate points of views that address the query from different perspectives. WikiContradict~\citep{hou2024wikicontradict} leverages Wikipedia tags to identify contradictions and ask human annotators to write questions that reflect the contradiction between two paragraphs. QACC~\citep{liu-etal-2025-open} is a subset of unambiguous queries from AmbigQA~\citep{min-etal-2020-ambigqa}, where human annotators were asked to write the different answers from the search snippets and to select the correct answer. AmbigDocs~\citep{lee2024ambigdocs} focuses on ambiguous queries to evaluate the ability of models to distinguish between different entities sharing the same name. \citet{wang2025retrieval} focus on multiple types of knowledge conflicts by extending AmbigDocs with automatically generated misinformation and retrieval-induced noise. They also suggest evaluating RAG systems with varying numbers of retrieved sources, similarly to real-world retrieval distributions.

In contrast to the existing resources, \dataset{} is the first RAG dataset to include human annotation of the category of the knowledge conflict, based on our proposed taxonomy (\S\ref{sec:taxonomy}). As shown in our experiments~(\S\ref{sec:experiments_results}), this information is valuable for generating an appropriate response. Furthermore, \dataset{} includes a diverse set of queries and the relevant documents constitute a real-world scenario of RAG where LLMs are augmented with search results. Therefore, \dataset{} can serve as a broad evaluation benchmark to assess how models handle a wide spectrum of knowledge conflict in RAG scenarios.

\subsection{LLM Evaluation}
\label{subsec:llm_eval}

The evaluation of LLMs has become a subject of intense research interest, assessing various aspects of their outputs, including factuality with respect to world knowledge or to a given context~\citep[\emph{inter-alia}]{rashkin-etal-2023-measuring, min-etal-2023-factscore,tang-etal-2024-minicheck, song-etal-2024-veriscore,Cattan2024LocalizingFI,Ravichander2025HALoGENFL, Jacovi2025TheFG}, instruction-following~\citep{skopek-etal-2023-towards, liu-etal-2024-alignbench}, coherence~\citep{gomez-rodriguez-williams-2023-confederacy}, \emph{inter-alia}. 
We expand the evaluation scope and introduce an evaluation methodology that assesses not only whether LLMs resolve knowledge conflicts in RAG, but \textit{how} they do so, in alignment with human expectations.

\section{Conclusion}

This work highlights the critical role of the conflict type in Retrieval Augmented Generation (RAG). We hope \dataset{} will serve as a valuable resource for developing more robust RAG models. Beyond response generation, future work can explore how to leverage conflict type for other applications, such as enhancing the reasoning and decision-making capabilities of agentic LLMs.

\section*{Acknowledgments}

We thank Aviv Slobokdin for reviewing the paper draft and providing valuable feedback. We are also grateful to Gabriel Stanovsky, Roy Schwartz, Tu Vu, Adam Bloniarz, Corey Fry and Avigail Dabush for fruitful discussion at various stages of the project. We thank Siyi Liu for sharing the QACC dataset. We thank Michael Riley, Itay Laish and Dave Orr for reviewing the paper. Special thanks to Rebecca Galor for managing the annotation tasks, onboarding the annotators and providing them feedback along the process. Finally, we are grateful to all annotators that participated in the construction of \dataset{}.

\bibliography{anthology,tacl2021}
\bibliographystyle{acl_natbib}

\appendix

\section{Dataset}
\label{app:dataset}

Table~\ref{tab:datasources} presents the number of queries from each original datasource. 

\begin{table}[!t]
    \centering
    \resizebox{0.48\textwidth}{!}{
    \begin{tabular}{lc}
    \toprule
    Datasource     &  \# Instances \\
    \midrule
    ConflictingQA~\citep{wan-etal-2024-evidence}  & 162 \\
    SituatedQA Geo~\citep{zhang-choi-2021-situatedqa} & 105 \\
    SituatedQA Temp~\citep{zhang-choi-2021-situatedqa} & 41 \\
    FreshQA~\citep{vu-etal-2024-freshllms}  & 95\\
    QACC~\citep{Liu2024OpenDQ} & 55 \\
    \bottomrule
    \end{tabular}}
    \caption{Number of queries for each data source.}
    \label{tab:datasources}
\end{table}

\section{Prompts}
\label{app:prompts}

Figure~\ref{fig:conflict_type_prompt} shows the prompt we use to predict the conflict type given a query and its corresponding search results (\emph{Task 1} in \S\ref{sec:tasks}).

Figure~\ref{fig:vanilla_prompt} shows our vanilla prompt for generating the response with inline citations.
For the pipeline and the oracle approaches, we add a description of the conflict type to the Vanilla prompt. 

For the taxonomy-aware method, we provide the definition of each category, as in Figure~\ref{fig:conflict_type_prompt}, and prompt the model to predict the category, explain the decision and generate an appropriate response. 

\begin{figure*}[t]

\lstdefinestyle{promptStyle}
{
    basicstyle={\footnotesize\ttfamily},%
    xleftmargin=2.8em,%
    xrightmargin=1.5em,
    showstringspaces=false,
      showspaces=false,
        showtabs=false,
    tabsize=2,
    breaklines=true,
        flexiblecolumns=true,
        escapeinside={<@}{@>},
          breakatwhitespace=true
}

\newtcblisting{mylisting}[1]{
  enhanced,
  listing only,
  boxrule=0.8pt,
  sharp corners=downhill,
  top=0mm,
  bottom=0mm,
  left=2mm,
  right=0mm,
  boxsep=0mm,
  colframe=black,
  colback=white,
  listing options={
    style=#1
  }
}

\definecolor{instructionsColor}{rgb}{0.1, 0.5, 0.1}

\begin{mylisting}{promptStyle}
You are tasked with analyzing the search results provided for a given query and classifying the type of knowledge conflict present (if any).
Consider the query and the search results carefully, then classify the conflict into *one* of the following categories, using the descriptions and examples provided below.

1. <@\textbf{No Conflict:}@> The search results refer to the same concept and are in agreement. Differences are superficial, such as variations in detail or granularity.
    *   *Example:*
        *   *Query:* What is the meaning of the name Apoorva?
        *   *Search Results:* Unique, Quite new, Not seen before
2.  <@\textbf{Complementary Information:}@> The question is underspecified or allows for multiple valid perspectives or answers that do not contradict each other. All provided answers can be considered correct.
    *   *Example:*
        *   *Query:* Is public transportation faster than driving in cities?
        *   *Search Results:* Depends on the city and situation (e.g., rush hour vs. off-peak), specific routes matter, availability of parking is relevant.
3.  <@\textbf{Conflicting Opinions or Research Outcomes:}@> The query addresses a subjective or contentious topic, leading to genuinely opposing viewpoints or contradictory research findings.
    *   *Example:*
        *   *Query:* Is online learning as effective as traditional classroom learning?
        *   *Search Results:* Some sources argue "yes," citing flexibility and accessibility. Others argue "no," emphasizing the importance of in-person interaction.
4.  <@\textbf{Conflict Due to Outdated Information:}@> The question has a verifiable factual answer, but the search results contain conflicting information due to changes over time.
    *   *Example:*
        *   *Query:* Do Tesla and X Corp. have the same CEO?
        *   *Search Results:* Some articles say yes (dated before the change), others say no (dated after the change).
5.  <@\textbf{Conflict Due to Misinformation:}@> The question has a verifiable factual answer, but some of the search results contain factually incorrect or misleading information.
    *   *Example:*
        *   *Query:* What is the capital of Israel?
        *   *Search Results:* One source correctly states Jerusalem, while another incorrectly states Tel Aviv.


<@\textbf{Input:}@>
*   QUERY: {query}
*   SEARCH RESULTS: {search_results}

<@\textbf{Output:}@>
Please provide your response as a JSON object with the following fields:
*   <@\textbf{explanation:}@> (String) A brief explanation of why you chose this category.
*   <@\textbf{category:}@> (Integer) The number (1-5) corresponding to the category of the knowledge conflict.
\end{mylisting}

\caption{Prompt for predicting the conflict type.}
\label{fig:conflict_type_prompt}
\end{figure*}

\begin{figure*}[t]

\lstdefinestyle{promptStyle}
{
    basicstyle={\footnotesize\ttfamily},%
    xleftmargin=2.8em,%
    xrightmargin=1.5em,
    showstringspaces=false,
      showspaces=false,
        showtabs=false,
    tabsize=2,
    breaklines=true,
        flexiblecolumns=true,
        escapeinside={<@}{@>},
          breakatwhitespace=true
}

\newtcblisting{mylisting}[1]{
  enhanced,
  listing only,
  boxrule=0.8pt,
  sharp corners=downhill,
  top=0mm,
  bottom=0mm,
  left=2mm,
  right=0mm,
  boxsep=0mm,
  colframe=black,
  colback=white,
  listing options={
    style=#1
  }
}

\definecolor{instructionsColor}{rgb}{0.1, 0.5, 0.1}

\begin{mylisting}{promptStyle}
Write a high-quality and concise answer for the given question using only the provided search results. 
For each sentence, cite the corresponding sources using [1][2][3].

QUERY: {query}
SEARCH RESULTS: {search_results}
\end{mylisting}

\caption{Template prompt for Vanilla.}
\label{fig:vanilla_prompt}
\end{figure*}

\iftaclpubformat

\onecolumn

\fi

\end{document}